\documentclass[english,onecolumn]{article}
\usepackage{times}
\usepackage[T1]{fontenc}
\usepackage[utf8]{inputenc}
\usepackage{array}
\usepackage{verbatim}
\usepackage{float}
\usepackage{url}
\usepackage{amsmath}
\usepackage{amssymb}
\usepackage{graphicx}
\usepackage{epstopdf}
\usepackage{babel}
\usepackage{color}
\usepackage{hyperref}
\usepackage[numbered,framed]{mcode}

\usepackage[scale = 0.7]{geometry}

\newcommand{\R}{\mathbb{R}} 
\newcommand{\Rbb}{\mathbb{R}}

\DeclareMathOperator*{\argmin}{arg\,min}
\newcommand{\unlocbox}{UNLocBoX~}

\title{UNLocBoX \\ \vspace{2 mm} {\large A MATLAB convex optimization toolbox for proximal-splitting methods}}
\author{Nathanael Perraudin, Vassilis Kalofolias, David Shuman, Pierre Vandergheynst}
\date{October 2016}

\begin{document}
\maketitle


\begin{abstract}
Convex optimization is an essential tool for machine learning, as many of its problems can be formulated as minimization problems of specific objective functions. While there is a large variety of algorithms available to solve convex problems, we can argue that it becomes more and more important to focus on efficient, scalable methods that can deal with big data. When the objective function can be written as a sum of ``simple'' terms, proximal splitting methods are a good choice. \unlocbox is a MATLAB library that implements many of these methods, designed to solve convex optimization problems of the form 
$
\min_{x \in \mathbb{R}^N} \sum_{n=1}^K f_n(x).
$
It contains the most recent solvers such as FISTA, Douglas-Rachford, SDMM as well a primal dual techniques such as Chambolle-Pock and forward-backward-forward. It also includes an extensive list of common proximal operators that can be combined, allowing for a quick implementation of a large variety of convex problems.
\end{abstract}


\section{Introduction}

When solving a machine learning problem, very often we need to transform it first into an \textit{optimization problem} before being able to solve it. This amounts to expressing the cost of a possible solution $x$ as an objective function $f(x)$, and then try to minimize it in to achieve the best result $x^*$:
\begin{equation}\label{eq:prob1}
x^* = \arg\min_x f(x).
\end{equation}
Having an explicit objective function to minimize before designing an algorithm not only makes its solution easier, but also can make the goal clearer.

Many models lead to \textit{convex} objective functions. Convexity is a very desirable trait, as then we can always find a\footnote{This does not always imply that there is only one solution, for this we need \textit{strong convexity}, that is the case in most interesting problems.} solution $x^*$ of problem \eqref{eq:prob1}, at the \textit{global minimum} of the function $f(x)$. If the latter is not convex, we usually need to compromise with finding a \textit{local minimum} that is suboptimal with respect to the objective function, as it only approximately solves Problem \eqref{eq:prob1}.

Many problems in machine learning have an objective function that can be written as a sum of simpler ones:
\begin{equation}\label{eq:prob2}
x^* = \arg\min_{x\in\mathbb{R}^N}~\sum_{n=1}^Kf_n(x).
\end{equation}
This is for example the kind of objective function used for many maximum a posteriori (MAP) estimators, where each additional prior or constraint leads to a new term $f_i(x)$. We will see this example in a bit more detail in the next section. 

When the objective function has this structure, it is beneficial to take advantage of it, by using \textit{proximal splitting} optimization methods. As we explain in the next section, these methods solve simpler problems, based on each $f_i(x)$ separately, and use them as intermediate steps towards a global minimum of $f(x)$. Proximal splitting methods have the big advantage of being scalable (depending on $f_i$) especially due to this decoupling of the different parts of the objective function. 
%
%
This is especially important in machine learning applications, as (1) the number of variables is typically large and (2) accuracy is only relatively important. Therefore, recently \textit{we observe a paradigm shift from methods whose main objective is convergence in few iterations\footnote{A few years ago, the trend was to focus on convergence to large accuracy in few iterations, that made Newton-iteration-based methods (like interior-point methods) very popular. However, these methods have a high per-iteration complexity, typically quadratic in the number of variables, that makes them impractical for modern large-scale applications.} to methods whose main goal is a low per-iteration complexity}.

\subsection{UNLocBoX}
UNLocBoX is a convex optimization toolbox for solving problems in the form \eqref{eq:prob2} with MATLAB. It focuses especially on proximal splitting methods, and our goal is to keep its use \textit{simple}, and at the same time very efficient and suitable for solving \textit{large-scale problems}. UNLocBoX is to be used by both novice and advanced users of optimization. For the novice, it suffices to define the objective function as a sum of simpler ones, and a general solver will automatically choose the optimization framework that is more suitable for that specific problem. The advanced user, on the other hand, has a great level of control, as he can choose the solver he prefers from a wide variety of proximal splitting methods, and change the parameters according to his needs. This freedom of choice makes UNLocBoX also a great hands-on tool for \textit{learning} or practicing convex optimization.

UNLocBoX, however, is not to be used completely blindly as a black box. It requires a minimal knowledge from the user, that is (1) to be able to write the objective function as a sum of simpler ones, and (2) to identify if each of them is differentiable or not. For the differentiable he needs to provide the gradient. For the non-differentiable, he might need to provide a proximal operator (see next Section), if it does not already exist among the ones already implemented in the toolbox. For users that want to use a complete black box optimization solution, we refer them to CVX \cite{grant2008cvx}. The reason for using UNLocBoX instead is that it is much more efficient, and scalable for a much larger number of variables, as it exploits the structure of the objective function to efficiently solve each sub-problem in an iterative fashion.

The documentation of the UNLocBoX is complete, meaning that every single function is documented. Even if it is not perfect, it has improved significantly with time and we expect it to continue to do so. You can find it online at \url{https://lst2.epfl.ch/unlocbox/doc}. Everything is also periodically updated into a \href{https://lst2.epfl.ch/unlocbox/notes/unlocbox-note-003.pdf}{big PDF file}.


The design of UNLocBoX was largely inspired by the \textbf{LTFAT} toolbox \cite{ltfatnote030}. The authors are jointly developing mat2doc a documentation system that allows to generate documentation from source files.

\section{Proximal splitting}
To understand the idea behind proximal-splitting methods, let us take the simple example of linear regression from noisy linear observations $y\in\mathbb{R}^M$. Assuming that $y\approx Ax$ with Gaussian noise, we want to infer $x\in\mathbb{R}^N$ using a maximum a-posteriori (MAP) estimator. If further assume a Gaussian prior on $x$, the MAP estimator is given by
\begin{equation}
x^* = \arg\min_{x\in\mathbb{R}^N} \| A x - y \|_2^2 + \gamma\| x \|_2^2.
\end{equation}
This problem is not particularly difficult: both parts of the objective function $f(x)$ are differentiable, and we know that the solution of our problem lies at the point where the gradient of $x$ is zero: $\nabla_x f(x) = 0$. This condition leads to a linear system of equations, but since we are interested in scalable solutions we would not try to directly solve the system (that costs $\mathcal{O}(N^3)$). A more scalable solution is to use the gradient descent method, which has a cost of only $\mathcal{O}(NM)$ per iteration:
\begin{equation}\label{eq:grad_descent}
x^i \leftarrow x^{i-1}-\lambda\nabla_x f|_{x=x^{i-1}},
\end{equation}
where $f\left|_{x=x^{i-1}}\right.$ is the gradient of $f$ with respect to $x$, computed at the point $x=x^{i-1}$, and $\lambda$ is the step-size. 

Taking into account the splitting of $f(x)=f_1(x)+f_2(x)$ with $f_1(x)=\| A x - y \|_2^2$ and $f_2(x)=\|x\|_2^2$, we can rewrite the iteration \eqref{eq:grad_descent} as
\begin{align}
x^i & \leftarrow x^{i-1}-\lambda \nabla_x f_1|_{x=x^{i-1}},\\
x^i & \leftarrow x^{i}-\lambda \nabla_x f_2|_{x=x^{i-1}},
\end{align}
or even approximate it with
\begin{align}
x^i & \leftarrow x^{i-1}-\lambda \nabla_x f_1|_{x=x^{i-1}},\\
x^i & \leftarrow x^{i}-\lambda \nabla_x f_2|_{x=x^{i}},
\end{align}
where the second step uses the solution $x^i$ computed after the first step. Both the above iterative schemes are guaranteed to converge to the minimum, as long as $\lambda$ is small enough. Obviously, when making a step that reduces the value of $f_1(x)$, we might increase the value of $f_2(x)$. By keeping $\lambda$ small, the effect of reducing the first is stronger than the effect of increasing the second (as we choose the steepest descend direction for $f_1$ using its gradient).

The idea of proximal splitting is to generalize this iteration for non-differentiable functions. Suppose for example that instead of a Gaussian prior we have a Laplace prior on $x$. Then the problem we need to solve is 
\begin{equation}
x^* = \arg\min_{x\in\mathbb{R}^N} \| A x - y \|_2^2 + \gamma\| x \|_1.
\end{equation}
As the second term is not differentiable, we replace the corresponding gradient descent step with a \textit{proximal step}, that is obtained as the solution of
\begin{equation}\label{eq:prox_l1}
x^i \leftarrow \arg\min_{x\in\mathbb{R}^N} \gamma\|x\|_1 + \frac{1}{2}\|x - x^{i-1}\|^2,
\end{equation}
where $\gamma$ again plays the role of the step size. When $\gamma$ is small, solving the above problem will obtain a solution very close to the one of the previous step, but for which the norm-1 of $x$ is still smaller. Alternating between a gradient step for $f_1$ and a proximal step for $f_2$ is the well-known \textit{proximal gradient or forward-backward} algorithm, also known as the \textit{iterative soft thresholding algorithm} \cite{gabay1983chapter}-\nocite{tseng1991applications} or \textit{ISTA} for this specific choice of $f_2$. Simply choosing wisely the step sizes can accelerate the convergence rate of the algorithm, leading to the \textit{Fast ISTA} or \textit{FISTA} \cite{beck2009fast}. 

Note that the complexity of solving each sub-problem is $\mathcal{O}(N)$, therefore the overall complexity scales only linearly to the number of variables $N$. This is of huge importance for many machine learning problems where there is a large number of variables, and makes proximal splitting techniques more usable in practice than for example second order methods.

In UNLocBoX we implement many algorithms like the ones above, along with the Douglas-Rachford algorithm \cite{douglas1956numerical,lions1979splitting}-\nocite{lions1979splitting,eckstein1992douglas,combettes2004solving}\cite{combettes2007douglas}, capable of solving the problem for two non-differentiable terms, two primal-dual-based algorithms \cite{komodakis2015playing} that deal with two non-differentiable and a differentiable term, and other generalized algorithms that can handle more than 2 terms in $f(x)$. For an overview of proximal splitting methods we refer the reader to the work of \cite{combettes2005signal}, and for primal-dual algorithms to the review of \cite{komodakis2015playing}.


\subsection{The proximal operator}


In general, a proximal operator of a lower semi-continuous convex function $f_i$ from $\R^N$ to $\left(-\infty,+\infty\right]$ is defined as
\begin{align}
\label{def_prox}
{\rm prox}_{f} (x) := \argmin_{y \in \R^N} \left\{ \frac{1}{2} \|x-y\|_2^2 + f(y)\right\}.
\end{align}
%
%
%
%
%
Proximal operators have many interesting properties that make them very useful in iterative methods.
For example, note that the objective function is strictly convex, therefore the optimization problem \eqref{def_prox} has a unique solution, making the proximal operator well-defined.

A very important property of the proximal operator is that for an input $x_i$ it provides an output $x_{i+1}$ that obtains a smaller value of $f$:
\[x_{i+1} = {\rm prox}_{f} (x_i)  \Rightarrow f(x_{i+1}) \leq f(x_{i}),\]
with equality only if
\[x_i = x_{i+1} = \arg\min_{x\in\mathbb{R}^N}f(x).\]
Therefore, the iteration $x_{i+1} = {\rm prox}_{f} (x_i)$ leads to the minimum of $f(x)$.
Proximal splitting methods can be seen as methods that generalize this iteration, alternating between the proximal operators of different terms $f_i$ of $f$.

The success of proximal splitting methods for scalable applications is based on the fact that for many convex functions $f(x)$, the corresponding proximal operator \eqref{def_prox} can be solved using only $\mathcal{O}(N)$ operations. For example, solving \eqref{eq:prox_l1} for $f(x) = \|x\|_1$ is done by elementwise \textit{soft-thresholding}:
\begin{equation}
\text{soft}_\gamma(\alpha) = 
\begin{cases}
\max(0, \alpha-\gamma), &\text{if}~~\alpha\geq0,\\
\min(0, \alpha+\gamma), &\text{if}~~\alpha<0.
\end{cases}
\end{equation}
In the great review \cite{combettes2011proximal} there is a list of many convex terms along with their proximal operators, while it is usually easy to define our own.

\section{Structure of the UNLocBoX}

The most important function of the toolbox is \verb?solvep?, as it automatically chooses a specific solver for a given objective function given as a sum of convex terms.
Additionally, the toolbox mainly contains four groups of functions:
\begin{enumerate}
\item \textbf{Solvers.} (See Section~\ref{sec:solvers}) They form the core of the toolbox and are usually called by \verb?solvep?. The toolbox includes most of the recent techniques, like forward-backward (FISTA), Douglas-Rachford, PPXA, SDMM, as well as many primal-dual techniques.
\item \textbf{Proximal operators.} (See Section~\ref{sec:proximal_unloc}) The proximal operators of the most common functions help the user quickly solve standard problems. The toolbox also includes many projection operators that are needed for adding hard constraints.
\item \textbf{Demonstration files.} They show how to start with the toolbox.  
\item \textbf{Utility functions.} A set of small functions that are useful for various problems of the toolbox, including functions that could easily be part of standard MATLAB.
\end{enumerate}
An optimization problem is solved in two steps. First, the user needs to define the parts of the objective function (Sections \ref{sec:diff_functions} and \ref{sec:proximal_unloc}) and then choose a solver (Section \ref{sec:solvers}) or let the software automatically select one.

In Problem \eqref{eq:prob2}, each function $f_k(x)$ is modeled as a structure with special fields: \verb?eval?, \verb?prox?, \verb?grad?, \verb?beta?. For a function \verb?f?, the field \verb?f.eval? is a MATLAB function handle that takes as input the optimization variables $x$ and returns the value $f(x)$. In MATLAB, write: 
\begin{lstlisting}
    f.eval = @(x) eval_f(x)
\end{lstlisting}
where \verb?eval_f(x)? returns the scalar value of $ f (x) $.

\subsection{Defining differentiable functions}\label{sec:diff_functions}

If the function $f$ is differentiable (and we can compute the gradient), we have to specify the gradient in the field  \verb?f.grad? as a handle of a function that takes as input the optimization variables $x$ and returns $\nabla f(x)$. In MATLAB, write: 
\begin{lstlisting}
    f.grad = @(x) grad_f(x)
\end{lstlisting}
where \verb?grad_f(x)? returns the value of $\nabla  f (x) $. For differentiable functions, the field \verb?beta? needs also to be specified. It contains an upper bound on the Lipschitz constant of the gradient.
A function $f$ has a $\beta$-Lipschitz-continuous gradient $\nabla f$ if
\begin{equation}
\| \nabla f(x) - \nabla f(y) \|_2 \leq  \beta  \|x-y\|_2 \, \hspace{5mm} \forall x,y \in \R^N,
\end{equation}
where $\beta > 0$. In other words, the Lipschitz constant is an upper bound of the norm of the gradient operator.
As an example, to define the function $f(x) = 5 \| Ax - y\|_2^2$, you would write:
\begin{lstlisting}
    f.eval = @(x) 5 * norm(A*x - y)^2;
    f.grad = @(x) 2 * 5 * A'*(A*x - y); 
    f.beta = 2 * 5 * norm(A)^2;
\end{lstlisting}
Note that we use the operator norm (norm-2 of a matrix) of $A$, and the fact that $\|A^\top A\| = \|A\|^2$. Because the gradient can be written as $10A^\top(Ax - y) = (10A^\top A)x + \text{constant}(x)$, the Lipschitz constant is equal to $\|10A^\top A\|$.

\subsection{Defining proximal operators} \label{sec:proximal_unloc}

If the function $f$ is not differentiable, it has to be minimized using its proximal operator. For a lower semi-continuous function $f$, they are defined as:
\begin{align*}
{\rm prox}_{\lambda f} (x) := \argmin_{y \in \R^N} \left( \frac{1}{2} \|x-y\|_2^2 + \lambda f(y)\right).
\end{align*}
In order to facilitate the definition of functions and to allow for fast implementations, the UNLocBoX includes a variety of proximal operators. The proximal operators of \unlocbox are defined as MATLAB functions that take three input parameters: \texttt{x, lambda, param}. First, \texttt{x} is the initial signal, that is, the signal in the current iteration, before applying the proximal operator. Then \texttt{lambda} is the weight of the objective function, multiplied by a step-size needed by each algorithm. Finally, \texttt{param} is a MATLAB structure that contains a set of optional parameters.

In this case, the field  \verb?f.prox? is another function handle that takes as input a vector $x$ along with a positive real number $\tau$  and returns ${\rm prox}_{\tau f}(x)$. The parameter $\tau$ is usually decomposed into $\gamma \cdot c$ where $\gamma$ is the step-size needed by the algorithm and $c$ is the weight of the function. To define a function $c \cdot f(x)$, in MATLAB, we write: 
\begin{lstlisting}
    f1.prox = @(x, T) prox_f1(x, c*T)
\end{lstlisting}
where \verb?prox_f1(x, T)? can be a standard MATLAB function that solves the problem ${\rm prox}_{T f_1} (x) $ given in equation \eqref{def_prox}. Note that we use this form of input so that the solver is free to apply the step-size it needs, that is equivalent to multiplying $f(x)$ by a constant.

The most typical non-differentiable function used in optimization is the $\ell_1$-norm. It is used as a relaxation for the $\ell_0$ norm and thus favor sparsity. Fox example the function $f(x) = 7 \|x\|_1 $ can be implemented as:
\begin{lstlisting}
    f.eval = @(x) 7 * norm(x, 1);
    f.prox = @(x, T) prox_l1(x, 7*T); 
\end{lstlisting}
In many applications, the signal is assumed to be sparse within a special set of atoms $W$. In this case, the basis is specified in the optional argument of the function. As an example, the function $f(x) = 3 \|W x\|_1$ is encoded as:
\begin{lstlisting}
    f.eval = @(x) 3 * norm(x,1);
    param_l1.A = @(x) W * x;
    param_l1.At = @(x) W' * x;
    param_l1.tight = 0;
    param_l1.nu = norm(W)^2;
    f.prox = @(x, T) prox_l1( x, 3*T, param_l1 ); 
\end{lstlisting}
In this special example, we assume that $W$ is a matrix. The argument \verb?param_l1.nu? is very important. It is an upper bound on the squared norm of the operator:
$$
\sqrt{\nu} \geq \max_{x\neq 0} \frac{\|Wx\|_2}{\|x\|_2} = \lambda_{\max} (W).
$$
Note that if $W W^* = I$, meaning that the operator $W^*$ is \textit{tight}, the parameter \verb?param_l1.tight? can be set to $1$ so that simpler computations can be used.
In the non-tight case, the algorithm will perform inner iterations. As inner iterations leads to slower optimization, it is recommended avoiding this case. To do so, primal-dual solvers can be used. Their presentation is beyond the scope of this paper. 


As a general rule, the user defines a function with its gradient whenever it is possible and keep the proximal form for non-differentiable functions. This usually allows the use of more efficient solvers.

The UNLocBoX includes the proximal operators of the most common norms (a non-exhaustive list is available in Table~\ref{norms_presentation}).
\begin{table} [!h]
\centering
\begin{tabular}{|l|p{6cm}|l|}
    \hline
    \textbf{Name}  & \textbf{Discrete definition} & \textbf{Input type} \\
    \hline
    \hline
$\ell_1$ norm &  $\|x\|_1=\sum_{n=1}^N |x_n|$ & Vector \\ \hline
$\ell_2$ norm (squared) &  $\|x\|_2= \sum_{n=1}^N |x_n|^2$ & Vector  \\ \hline
$\ell_{12}$ mixed norm &  $\|x\|_{12} = \sum_{g \in G} \|x_g\|_2,$ where $x_g \subset x$     are different groups of elements& Vector or matrix\\ \hline
$\ell_{TV}$ total variation norm  & $\|x\|_{TV}=\sum_{n} \|\nabla x(n) \|_2 = \sum_{n} \sqrt{\nabla_1 x(n) + \nabla_2 x(n)} $ (2 dimensional) & Matrix \\ \hline
$\ell_*$ nuclear norm & $\|x\|_*=\sum_{i=1}^{\min\{m,n\}}\sigma_i$, where $\sigma_i$ are the singular values of $x$ & Matrix  \\ \hline
\end{tabular}
\caption{\label{norms_presentation} Presentation of the norms}
\end{table}

\unlocbox solvers and proximal operators are not linked to a particular input size. As a result, the optimization variables can be in a vector, matrix, or tensor form. However, the user should make sure that the returned arguments of the gradient and of the proximal operator are of the same size as the one of the optimization variable $x$.

\subsubsection*{Adding constraints}
In order to restrict the set of admissible solutions to a convex set $\mathcal C\subset\Rbb^L$, i.e. find the optimal solution to \eqref{eq:prob1} considering only solutions in $\mathcal C$, we can use projection operators instead of proximal operators. Indeed, proximal operators are generalizations of projections. For any non-empty, closed and convex set $\mathcal C\subset\mathbb{R}^{L}$, the \emph{indicator function} of $\mathcal C$ is defined as
\begin{equation}
i_{\mathcal C}:\mathbb{R}^{L}\rightarrow\{0,+\infty\}:x\mapsto\begin{cases}
0,\hspace{0.25cm} & \text{if}\hspace{0.25cm}x\in \mathcal C\\
+\infty\hspace{0.25cm} & \text{otherwise.}
\end{cases},\label{eq:indfun}
\end{equation}
The corresponding proximal operator is given by the projection onto the set $\mathcal{C}$:
\begin{align*}
P_\mathcal{C}(y) & = \mathop{\operatorname{arg~min}}\limits _{x\in\mathbb{R}^{L}}\left\{ \frac{1}{2}\|y-x\|_{2}^{2}+ i_\mathcal{C}(x)\right\}\\
 & = \mathop{\operatorname{arg~min}}\limits _{x\in\mathcal{C}}\left\{ \|y-x\|_{2}^{2} \right\}
\end{align*}
Such constraints are useful, for example, when we want the solution to be restricted to satisfy a linear equation or inequality. 

For consistency, the projection operators also take $3$ arguments as input. However the second argument is simply ignored by the solver, as an indicator function is invariant to multiplications by any positive number. To implement the constraint $\| x \|_2\leq2$, one would simply write
\begin{lstlisting}
    f.eval = @(x) eps;
    param_b2.epsilon = 2
    f.prox = @(x, T) prox_b2(x, T, param_b2);
\end{lstlisting}
For constraints, we do not specify a particular objective function. Theoretically, it should be infinite when the constraint is not satisfied. In practice, infinite is not manageable for a computer and it is simply better to ignore this case.

\subsection{Solvers}\label{sec:solvers}

The \unlocbox solvers are categorized them into 2 different groups. First, there are specific solvers that minimize only two functions (Forward backward, Douglas-Rachford, ADMM, forward-backward-forward, Chambolle-Pock,...). Those are usually more efficient. Second, there are general solvers that are more general but less efficient (Generalized forward backward, PPXA, SDMM,...). Not all possible solvers are included into the UNLocBoX. However the toolbox offers a general framework, where the user can add his own solvers.

In general a solver takes three kinds of inputs: an initialization point $x_0$, the functions to be minimized and an optional structure of parameters. As a rule into the UNLocBoX, we use the following convention. The initialization point is always the first argument (except for SDMM), then comes the functions and finally the optional structure of parameters. 

\paragraph{Selection of a solver}
The UNLocBoX contains a general solving function called \verb?solvep? able to select a good solver for in most of the cases. Additionally, if some problem functions are differentiable, it computes an optimal time-step automatically. To minimize the sum of the functions \verb?f1, f2, f3?, we write in MATLAB:
\begin{lstlisting}
    sol = solvep(x_0, {f1, f2, f3}, param);
\end{lstlisting}
The solver can also be selected manually with:
\begin{lstlisting}
    sol = forward_backward(x_0, f1, f2, param);
\end{lstlisting}
Optional solver parameters are all contained into the optional structure \verb?param?. Different functions might need different parameters, but most of them are common. Table~\ref{tab:optional-parameters} presents a summary of the most important one. The solvers return 2 arguments, the solution, i.e. the minimizer of the problem, and a structure of general information that is detailed in Table~\ref{tab:return_info}.

\paragraph{Selection of the time-step} The UNLocBoX is automatically selecting the best time step based on the Lipschitz constant of the differentiable functions. However, if the problem contains only non-differentiable function, it may be necessary to introduce it manually in the optional field \verb?param.gamma?. In this case, the time-step defines the compromise between convergence speed and accuracy, lower value leading toward more accurate solutions.

\section{Example}
Let us suppose that we have a noisy image of dimension $n\times n$ with missing pixels. Our goal is to find the closest image to the original one. We assume that the positions of the missing pixels are known. We additionally assume the image to be composed of well-delimited patches of colors, i.e. to have a sparse gradient. Finally, we suppose that known pixels are subject to some Gaussian noise with a variance of $\epsilon$.
\begin{figure}[!ht]
\begin{center}
\includegraphics[width=25ex]{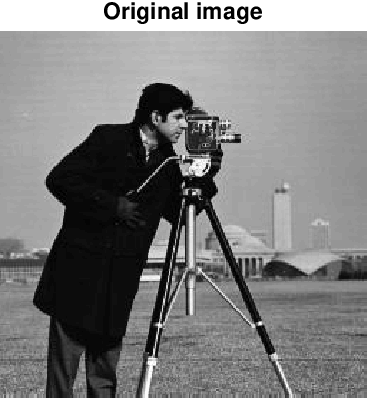}
\includegraphics[width=25ex]{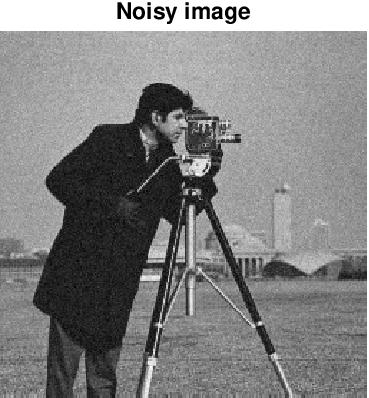}
\includegraphics[width=25ex]{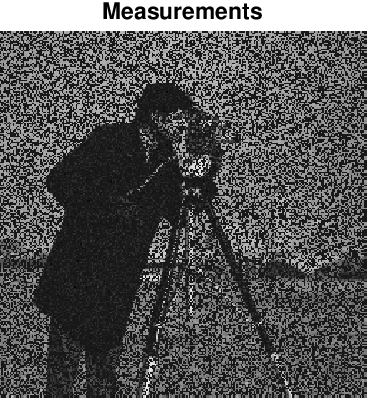}
\end{center}
\caption{This figure shows the image chosen for this example: the cameraman.}
\end{figure}

At this point, the problem can be expressed in a mathematical form. We will simulate the masking operation by a mask $A$. This leads to the generative process:
\begin{equation*}
y = Ax + n
\end{equation*}
where $x$ is the vectorized image we want to recover, $y$ are the observe noisy pixels and $n$ a vector encoding the noise. If we assume the noise to be Gaussian i.i.d, we can insert this generative model into our problem in form of a constraint:
\begin{equation*}
\mathcal{C} = \{x \in \Rbb^{n^2} \hspace{0.5cm} | \hspace{0.5cm} \| Ax - y \|_2 \leq \epsilon \}
\end{equation*}
Note that $\epsilon$ can be chosen equal to $0$ to exactly satisfy the constraint $Ax=y$ in the noiseless case.
In our case, as the measurements are noisy, we set $\epsilon$ to the standard deviation of the noise multiplied by $\sqrt{\#y}$, where $\#y$ is the number of elements of $y$. When the noise level is not known, this parameter has to be adjusted manually or using hyper-parameter selection techniques.

As we assume the image to have a sparse gradient, we add a total variation term in the optimization problem and we get:
\begin{equation*}
\argmin_x \|x\|_{TV} \hspace{1cm} \text{subject}\hspace{0.25cm}  \text{to}\hspace{1cm} \|Ax-y\|_2 \leq \epsilon \hspace{1cm} \text{(Problem I)},
\end{equation*}
where $ \|x\|_{TV} =\| \sqrt{\nabla_1x + \nabla_2 x} \|_1 $ is traditionally referred as the total variation norm\footnote{In fact it is only a semi norm as it does not satisfy the separability property.}. $\epsilon$ can also be seen as a parameter that tunes the confidence to the measurements. This is not the only way to define the problem. We could also write:
\begin{equation*}
\argmin_x \|Ax - y\|_2^2 + \lambda \|x\|_{TV} \hspace{1cm} \text{(Problem II)}
\end{equation*}
with the first term playing the role of a data fidelity term and the second a prior assumption on the signal. $\lambda$ adjusts the tradeoff between measurement fidelity and prior assumption. We call it the \emph{regularization parameter}. The smaller it is, the more we trust the measurements and vice-versa. $\epsilon$ play a similar role as $\lambda$.
Note that there exists a bijective mapping between the parameters $\lambda$ and $\epsilon$, leading to the same solution. The mapping function is not trivial to determine. Choosing between one or the other problem will affect the solvers and the convergence rate.
Note also that in Problem II we prefer to use the squared $\ell_2$ norm, as it is directly differentiable, including at $0$. This choice does not affect the range of solutions, only the mapping between $\lambda$ and $\epsilon$.

\paragraph{Solving Problem I}
The UNLocBoX solvers take as input functions with their proximal operator or with their gradient. In the toolbox, functions are modeled with structure objects containing special fields. One field contains an operator to evaluate the function and the other allows to compute either the gradient (in case of differentiable function) or the proximal operator ( in case of non-differentiable functions). In this example, we need to provide two functions:
\begin{itemize}
\item $f_1(x)=\|x\|_{TV}$, the total variation norm has a proximal operator defined as:
\begin{equation*}
prox_{f1,\gamma} (z) = arg \min_{x} \frac{1}{2} \|x-z\|_2^2  +  \gamma \|z\|_{TV}.
\end{equation*}
In MATLAB this function is defined with:
\begin{lstlisting}
    param_tv.verbose = 1;
    param_tv.maxit = 50;
    f1.prox = @(x,T) prox_tv(x , T, param_tv);
    f1.eval = @(x) norm_tv(x);
\end{lstlisting}
This function is a structure with two fields. First, \emph{f1.prox} is an operator taking as input $x$ and $T$ and evaluating the proximal operator of the function ($T$ plays the role of $\gamma$ is the equation above). Second, and sometime optional, \emph{f1.eval} is also an operator evaluating the function at $x$. 

The proximal operator of the TV norm is already implemented in the UNLocBoX by the function \texttt{prox\_tv}. We tune it by setting the maximum number of iterations and a verbosity level. Other parameters are also available (see documentation \url{https://lts2.epfl.ch/unlocbox/doc.php}).
\begin{itemize}
\item \emph{param\_tv.verbose} selects the display level (0 no log, 1 summary at
convergence and 2 display all steps).
\item \emph{param\_tv.maxit} defines the maximum number of iterations.
\end{itemize}

\item $f_2$ is the indicator function of the set S defines by $\|Ax-y\|_2 <\epsilon$. Its proximal operator of $f_2$ as
\begin{equation*}
prox_{f2,\gamma} (z) = arg \min_{x} \frac{1}{2} \|x-z\|_2^2   + i_S(x) ,
\end{equation*}
with $i_S(x)$ is zero if $x$ is in the set $S$ and infinite otherwise. This previous problem has an identical solution as:
\begin{equation*}
arg \min_{z} \|x - z\|_2^2   \hspace{1cm} \text{subject} \hspace{0.25cm} \text{to} \hspace{1cm} \|Az-by\|_2 \leq \epsilon
\end{equation*}
which is simply a projection on the B2-ball. In MATLAB, we write:
\begin{lstlisting}
    param_proj.epsilon = epsilon;
    param_proj.A = A;
    param_proj.At = A;
    param_proj.y = y;
    f2.prox = @(x, T) proj_b2(x, T, param_proj);
    f2.eval = @(x) eps;
\end{lstlisting}
The \emph{prox} field of \emph{f2} is in that case the operator computing the projection. Since we suppose that the constraint is satisfied, the value of the indicator function is $0$. For implementation reasons, it is better to set the value of the operator \emph{f2.eval} to \emph{eps} than to $0$. Note that this hypothesis could lead to strange evolution of the objective function as the constraint is not always satisfied during the optimization process. Here the parameter \emph{A} and \emph{At} are mandatory. Note that \emph{A} = \emph{At}, since the masking operator can be performed by a diagonal matrix containing 1's for observed pixels and 0's for hidden pixels.
\end{itemize}

At this point, a solver needs to be selected. The UNLocBoX includes a universal solving function \texttt{solvep} able to select a solver for your problem. The automatic choice might not be optimal as some solvers are optimized for specific problems. In this example, we present two of them \texttt{forward\_backward} and \texttt{douglas\_rachford}. Both of them take as input two functions (they have generalizations taking more functions), a starting point and some optional parameters.

\begin{figure}[!ht]
\begin{center}
\includegraphics[width=25ex]{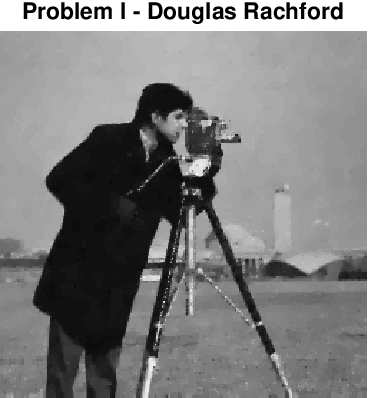}
\end{center}
\caption{\label{fig:sol prob 1}This figure shows the reconstructed image by solving Problem I using Douglas Rachford algorithm.}
\end{figure}

In our problem, neither function is  differentiable (for some points of the domain), leading to the impossibility to compute the gradient. In that case, solvers (such as forward backward) using gradient descent cannot be used. As a consequence, we will use Douglas-Rachford instead. In MATLAB, we write:
\begin{lstlisting}
    param.verbose = 1;
    param.maxit = 100;
    param.tol = 10e-5;
    param.gamma = 1;
    sol = douglas_rachford(y, f1, f2, param);
\end{lstlisting}
Note that if the last line could be replaced with 
\begin{lstlisting}
    sol = solvep(y, {f1, f2}, param);
\end{lstlisting}
the function \texttt{solvep} would automatically select Douglas-Rachford as a solver.
\begin{itemize}
\item{\emph{param.verbose} selects the display level (0 no log, 1 summary at}
convergence and 2 display all steps).
\item \emph{param.maxit} defines the maximum number of iterations.
\item \emph{param.tol} is stopping criteria for the loop. The algorithm stops if
\begin{equation*}
\frac{  n(t) - n(t-1) }{ n(t)} < tol,
\end{equation*}
where  $n(t)$ is the objective function at iteration $t$
\item \emph{param.gamma} defines the step-size. When only non-differentiable functions are minimized, it is a compromise between convergence speed and precision. When differentiable functions are minimized, a too large step-size \emph{gamma} may lead to the divergence of the algorithm. In this case, the user can let the UNLocBoX automatically set the step-size.
\end{itemize}
The solution is displayed in figure \ref{fig:sol prob 1}

\begin{figure}
\begin{center}
\includegraphics[width=25ex]{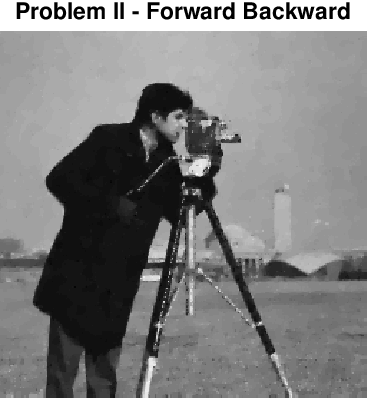}
\includegraphics[width=25ex]{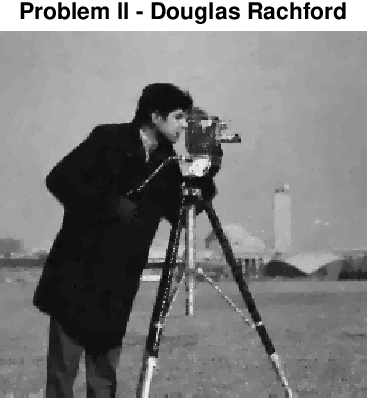}
\end{center}
\caption{\label{fig:sol prob 2} This figure shows the reconstructed image by solving problem II.}
\end{figure}

\paragraph{Solving Problem II}
Solving Problem II instead of Problem I can be done with a small modification of the previous code. First we define another function as follows:
\begin{lstlisting}
    param_l2.A = A;
    param_l2.At = A;
    param_l2.y = y;
    param_l2.verbose = 1;
    f3.prox = @(x,T) prox_l2( x, lambda * T, param_l2 );
    f3.grad = @(x) 2 * lambda * A(A(x) - y);
    f3.eval = @(x) lambda * norm(A(x) - y, 'fro')^2;
\end{lstlisting}
The squared $\ell_2$-norm is a differentiable function allowing the use of the forward backward solver. The gradient is defined in an additional field \texttt{grad}. In general, the user provides only the gradient or the proximal operator to a function. However, for this example we define both in order to compare two solvers. The proximal operator is used by the aforementioned Douglas-Rachford algorithm while the gradient is computed for the forward backward algorithm.
The solvers can be called by:
\begin{lstlisting}
    sol21 = forward_backward(y, f1, f3, param);
\end{lstlisting}
or:
\begin{lstlisting}
    sol22 = douglas_rachford(y, f3, f1, param);
\end{lstlisting}
These two solvers converge (up to numerical error) to the same solution. However, convergence speed might be different. As we perform only $100$ iterations with both of them, we do not obtain exactly the same result. Nonetheless, the differences cannot be perceived by eyes (see Figure~\ref{fig:sol prob 2}).

 \clearpage

\bibliographystyle{plain}
\bibliography{project}

\begin{thebibliography}{10}

\bibitem{beck2009fast}
A.~Beck and M.~Teboulle.
\newblock A fast iterative shrinkage-thresholding algorithm for linear inverse
  problems.
\newblock {\em SIAM Journal on Imaging Sciences}, 2(1):183--202, 2009.

\bibitem{combettes2004solving}
P.L. Combettes.
\newblock Solving monotone inclusions via compositions of nonexpansive averaged
  operators.
\newblock {\em Optimization}, 53(5-6), 2004.

\bibitem{combettes2007douglas}
P.L. Combettes and J.C. Pesquet.
\newblock A douglas--rachford splitting approach to nonsmooth convex
  variational signal recovery.
\newblock {\em Selected Topics in Signal Processing, IEEE Journal of},
  1(4):564--574, 2007.

\bibitem{combettes2011proximal}
P.L. Combettes and J.C. Pesquet.
\newblock Proximal splitting methods in signal processing.
\newblock {\em Fixed-Point Algorithms for Inverse Problems in Science and
  Engineering}, pages 185--212, 2011.

\bibitem{combettes2005signal}
P.L. Combettes and V.R. Wajs.
\newblock Signal recovery by proximal forward-backward splitting.
\newblock {\em Multiscale Modeling \& Simulation}, 4(4):1168--1200, 2005.

\bibitem{douglas1956numerical}
J.~Douglas and HH~Rachford.
\newblock On the numerical solution of heat conduction problems in two and
  three space variables.
\newblock {\em Transactions of the American mathematical Society},
  82(2):421--439, 1956.

\bibitem{eckstein1992douglas}
J.~Eckstein and D.P. Bertsekas.
\newblock On the douglas—rachford splitting method and the proximal point
  algorithm for maximal monotone operators.
\newblock {\em Mathematical Programming}, 55(1):293--318, 1992.

\bibitem{gabay1983chapter}
D.~Gabay.
\newblock Chapter ix applications of the method of multipliers to variational
  inequalities.
\newblock {\em Studies in mathematics and its applications}, 15:299--331, 1983.

\bibitem{grant2008cvx}
Michael Grant and Stephen Boyd.
\newblock Cvx: Matlab software for disciplined convex programming.

\bibitem{komodakis2015playing}
Nikos Komodakis and Jean-Christophe Pesquet.
\newblock Playing with duality: An overview of recent primal? dual approaches
  for solving large-scale optimization problems.
\newblock {\em IEEE Signal Processing Magazine}, 32(6):31--54, 2015.

\bibitem{lions1979splitting}
P.L. Lions and B.~Mercier.
\newblock Splitting algorithms for the sum of two nonlinear operators.
\newblock {\em SIAM Journal on Numerical Analysis}, 16(6):964--979, 1979.

\bibitem{ltfatnote030}
Zden\v{e}k Pr\r{u}\v{s}a, Peter~L. S{\o}ndergaard, Nicki Holighaus, Christoph
  Wiesmeyr, and Peter Balazs.
\newblock {The Large Time-Frequency Analysis Toolbox 2.0}.
\newblock In Mitsuko Aramaki, Olivier Derrien, Richard Kronland-Martinet, and
  S{\o}lvi Ystad, editors, {\em Sound, Music, and Motion}, Lecture Notes in
  Computer Science, pages 419--442. Springer International Publishing, 2014.

\bibitem{tseng1991applications}
P.~Tseng.
\newblock Applications of a splitting algorithm to decomposition in convex
  programming and variational inequalities.
\newblock {\em SIAM Journal on Control and Optimization}, 29(1):119--138, 1991.

\end{thebibliography}

\clearpage

\appendix

\section{Appendix: solver details}

\begin{table} [ht!]
\centering
\begin{tabular}{|p{4cm}|p{8cm}|}
   \hline
   \textbf{Field} & \textbf{Explanation} \\
   \hline
   \hline
  		\verb?tol? &  Tolerance used in the stopping criterion of the problem. \\ \hline
 		\verb?stopping_criterion? & Stopping criterion (see table~\ref{tab:stopping_crit}) \\ \hline
 		\verb?algo? & Selection of a specific solver \\ \hline
 		\verb?maxit? & Maximum number of iterations  \\ \hline
 		\verb?gamma? &  To manually specify the timestep. (To be used carefully)\\ \hline
 		\verb?verbose? & Verbosity level \\ \hline
 		\verb?debug_mode? &  Boolean to activate the computation of all internal variables. \\ \hline
\end{tabular}
\label{tab:optional-parameters}
\caption{Most common solver parameters as the fields of an optional structure.}
\end{table}

\begin{table} [ht!]
\centering
\begin{tabular}{|p{4cm}|p{8cm}|}
   \hline
   \textbf{Field} & \textbf{Explanation} \\
   \hline
   \hline
  		\verb?algo? &  Algorithm used \\ \hline
 		\verb?iter? & Number of iterations \\ \hline
 		\verb?time? & Time of execution of the function in sec.  \\ \hline
 		\verb?final_eval? & Final evaluation of the objective function  \\ \hline
 		\verb?crit? & Stopping criterion used see table \ref{tab:stopping_crit}  \\ \hline
 		\verb?rel_norm? & Relative norm at convergence   \\ \hline
\end{tabular}
\label{tab:return_info}
\caption{Information returned by the solver}
\end{table}   

\begin{table} [ht!]
\centering
\begin{tabular}{|p{4cm}|p{8cm}|}
   \hline
   \textbf{Criterion} & \textbf{Explanation} \\
   \hline
   \hline
   \verb?TOL_EPS? & Tolerance achieved \\ \hline
   \verb?ABS_TOL? & Objective function below the tolerance \\ \hline
   \verb?MAX_IT? & Maximum number of iterations\\ \hline
   \verb?USER? & Stop by the user "ctrl + D" in the command window.\\ \hline
   \verb?--? & Other\\
     \hline
\end{tabular}
\label{tab:stopping_crit}
\caption{Stopping criteria}
\end{table}   

\clearpage

\section{Appendix: example's code}
 
\begin{lstlisting}
%% Initialisation
clear;
close all;

% Loading toolbox
init_unlocbox;

verbose = 2; % verbosity level


%% Load an image

% Original image
im_original = cameraman; 

% Displaying original image
imagesc_gray(im_original, 1, 'Original image');  

%% Creation of the problem

sigma_noise = 10/255;
im_noisy = im_original + sigma_noise * randn(size(im_original));

% Create a matrix with randomly 50 % of zeros entry
p = 0.5;
matA = rand(size(im_original));
matA = (matA > (1-p));
% Define the operator
A = @(x) matA .* x;

% Masked image
y = A(im_noisy);

% Displaying the noisy image
imagesc_gray(im_noisy, 2, 'Noisy image');

% Displaying masked image
imagesc_gray(y, 3, 'Measurements');


%% Setting the proximity operator

lambda = 1;
% setting the function f1 (norm TV)
param_tv.verbose = verbose - 1;
param_tv.maxit = 100;
f1.prox = @(x, T) prox_tv(x, lambda*T, param_tv);
f1.eval = @(x) lambda * norm_tv(x);   

% setting the function f2 
param_proj.epsilon = sqrt(sigma_noise^2 * numel(im_original) * p);
param_proj.A = A;
param_proj.At = A;
param_proj.y = y;
param_proj.verbose = verbose - 1;
f2.prox = @(x, T) proj_b2(x, T, param_proj);
f2.eval = @(x) eps;


%% Solving problem I

% setting different parameters for the simulation
param_dg.verbose = verbose;    % display parameter
param_dg.maxit = 100;    % maximum number of iterations
param_dg.tol = 1e-5;    % tolerance to stop iterating
param_dg.gamma = 0.1 ;     % Convergence parameter
fig = figure(100);
param_dg.do_sol = @(x) plot_image(x, fig); % plotting plugin

% solving the problem with Douglas Rachord
param_dg.method = 'douglas_rachford';
sol = solvep(y, {f1, f2}, param_dg);

%% Displaying the result
imagesc_gray(sol, 4, 'Problem I - Douglas Rachford');


%% Defining the function for problem II

lambda = 0.05;
% setting the function f1 (norm TV)
param_tv.verbose = verbose-1;
param_tv.maxit = 50;
f1.prox = @(x, T) prox_tv(x, lambda * T, param_tv);
f1.eval = @(x) lambda * norm_tv(x);

% setting the function f3
f3.grad = @(x) 2 * A(A(x) - y);
f3.eval = @(x) norm(A(x) - y, 'fro')^2;
f3.beta = 2;

% To be able to use also Douglas Rachford
param_l2.A = A;
param_l2.At = A;
param_l2.y = y;
param_l2.verbose = verbose - 1;
param_l2.tightT = 1;
param_l2.pcg = 0;
param_l2.nu = 1;
f3.prox = @(x,T) prox_l2(x, T, param_l2);


%% Solving problem II (forward backward)
param_fw.verbose = verbose;    % display parameter
param_fw.maxit = 100;    % maximum number of iterations
param_fw.tol = 1e-5;    % tolerance to stop iterating
fig = figure(100);
param_fw.do_sol = @(x) plot_image(x, fig); % plotting plugin
param_fw.method = 'forward_backward';
sol21 = solvep(y, {f1, f3}, param_fw);
close(fig);

%% Displaying the result
imagesc_gray(sol21, 5, 'Problem II - Forward Backward' );   


%% Solving problem II (Douglas Rachford)
param_dg.method = 'douglas_rachford';
param_dg.gamma = 0.5 ;     % Convergence parameter
fig = figure(100);
param_dg.do_sol = @(x) plot_image(x, fig); % plotting plugin
sol22 = douglas_rachford(y, f3, f1, param_dg);
close(fig);

%% Displaying the result
imagesc_gray(sol22, 6, 'Problem II - Douglas Rachford');


%% Close the UNLcoBoX
close_unlocbox;

\end{lstlisting}

\end{document}